\documentclass[journal]{IEEEtai}

\usepackage[colorlinks,urlcolor=blue,linkcolor=blue,citecolor=blue]{hyperref}

\usepackage{color,array}

\newcolumntype{P}[1]{>{\centering\arraybackslash}p{#1}}
\newcolumntype{M}[1]{>{\centering\arraybackslash}m{#1}}

\usepackage{graphicx}
\usepackage{caption}
\usepackage{subcaption}
\usepackage{amsmath,amssymb,amsfonts}

\usepackage{algorithm}%
\usepackage{algorithmicx}%
\usepackage{algpseudocode}%

\usepackage{rotating}
\usepackage{booktabs}

\begin{document}

\title{OFAL: An Oracle-Free Active Learning Framework}

\author{Hadi Khorsand, Vahid Pourahmadi
\\
\IEEEmembership{{\small Department of Electronic Engineering, Amirkabir University of Technology,
Tehran, Iran}}

\thanks{Email addresses: Hadi Khorsand (hadi\_khorsand@aut.ac.ir), Vahid Pourahmadi (v.pourahmadi@aut.ac.ir).}
\thanks{Corresponding author is Hadi Khorsand.}
\thanks{Complete codes of the illustrate OFAL method available at https://github.com/Hadi-Khorsand/OFAL}}


\markboth{ }
{ }

\maketitle

\begin{abstract}
In the active learning paradigm, using an oracle to label data has always been a complex and expensive task, and with the emersion of large unlabeled data pools, it would be highly beneficial If we could achieve better results without relying on an oracle. This research introduces OFAL, an oracle-free active learning scheme that utilizes neural network uncertainty. OFAL uses the model's own uncertainty to transform highly confident unlabeled samples into informative uncertain samples. First, we start with separating and quantifying different parts of uncertainty and introduce Monte Carlo Dropouts as an approximation of the Bayesian Neural Network model. Secondly, by adding a variational autoencoder, we go on to generate new uncertain samples by stepping toward the uncertain part of latent space starting from a confidence seed sample. By generating these new informative samples, we can perform active learning and enhance the model's accuracy. Lastly, we try to compare and integrate our method with other widely used active learning sampling methods.
\end{abstract}



\section{Introduction}\label{sec1}
\label{sec:introduction}

Deep neural networks have been improved significantly in recent years and can perform highly complex applications. However, an ongoing problem in the deep learning paradigm is datasets. In order to train a neural network effectively, it needs large labeled datasets, but these datasets are not always available. Often, a large amount of data is not easily accessible, and additionally, it needs an oracle to label data, which is often too expensive. In today's world, because of the expansion of the internet and the rise of big data, the first problem has been solved in most cases, meaning we have access to a large amount of unlabeled data. Nevertheless, the cost and difficulty of labeling and the need for an oracle are still standing.

Therefore, the idea of active learning, which is achieving greater performance and accuracy with fewer labeled samples in the dataset, has always been an exciting topic in machine learning algorithms . Recent research in this paradigm developed a wide range of schemes and sampling methods for better performance \cite{settles2009active, fu2013survey}. Although active learning frameworks try to select the best possible samples from the unlabeled data pool to achieve the best results with the minimum number of labeled samples, there still is the need for an oracle. There is not always access to an oracle in real-world scenarios, which prevents us from using active learning methods. So it would be highly beneficial to improve the model without an expensive oracle to label new data.

The question before us is this: Can we reach higher neural network performance using the unlabeled data pool without an oracle?. In this research, we focused on the point that if the model does not perform well on some samples, the model has not yet seen samples from that sort of input or has not seen enough. This brings us to the concept of uncertainty in neural networks. The first idea in this scenario is to find the most uncertain samples from the unlabeled data pool. However, we need an oracle to provide labels to utilize them properly. Another idea is to use most certain samples to generate new uncertain samples. To do this, we introduce OFAL, an oracle-free active learning framework to generate and label new samples without needing an expensive oracle. OFAL starts with certain samples in the unlabeled data pool, which have low uncertainty and high prediction confidence from our model. Although these samples do not have labels, we can safely guess their actual label because of high model prediction confidence. Nevertheless, these samples are not informative for the model, meaning since we are already certain about them, adding them to our dataset does not improve the model's performance. Therefore, we use these samples to generate new informative uncertain samples by changing them to maximize their uncertainty but still carrying the same label.

We start with a brief review of the background and required literature in Section \ref{sec:2}, and go on to introduce OFAL and use uncertainty quantification to generate new samples in Section \ref{sec:3}. And in Section \ref{sec:4}, we will illustrate our result and integrate OFAL wth other sampling method schemes.

\begin{figure*}
	\centering
	\includegraphics[width=0.70\textwidth]{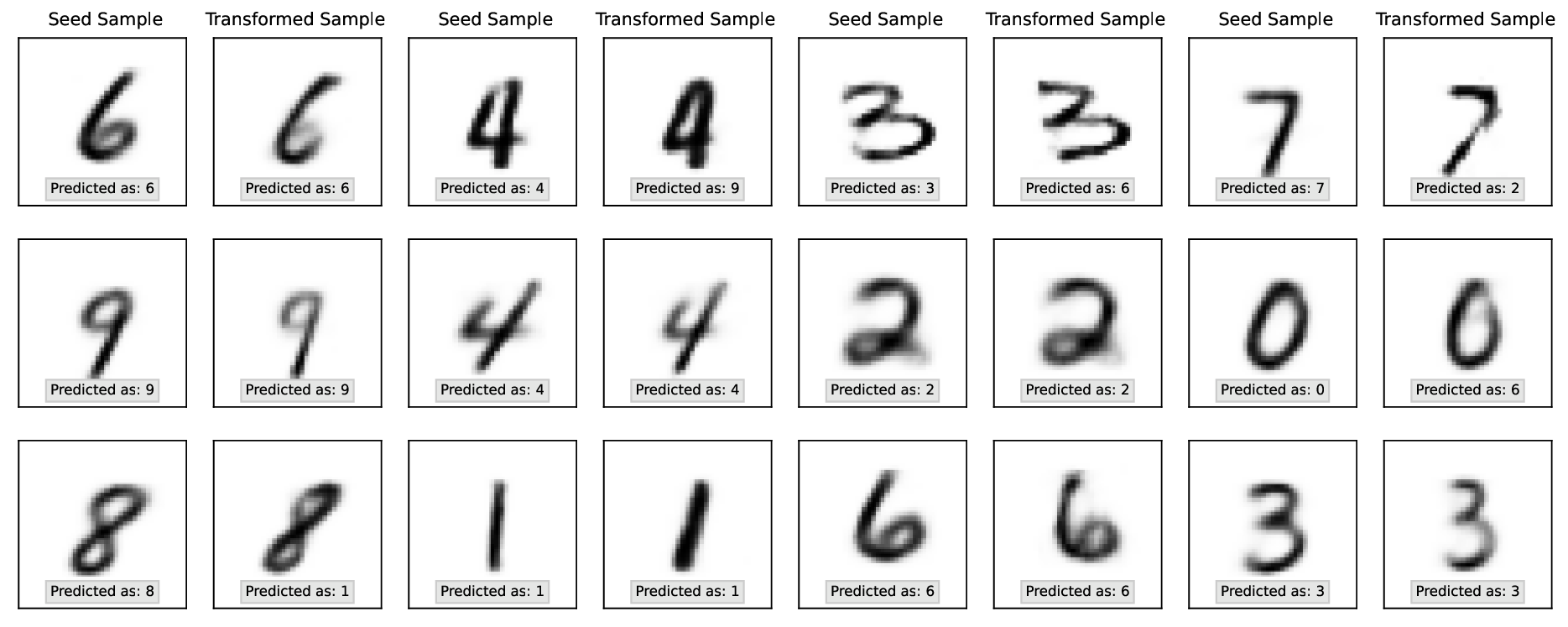}
	\caption{Example of samples generated by OFAL framework.}
	\label{fig:trans}
\end{figure*}

\section{Background}\label{sec2}
\label{sec:2}

We begin with a brief review of the required literature on Bayesian neural networks, uncertainty in neural networks, and active learning.

\subsection{Bayesian Deep Learning}

In general, a neural network model represents a function to produce output $\mathcal{Y}$ for a set of input $\mathcal{X}$, feasible by L layers and a set of weights and biases $w=[W_{l}, b_{l}]_{l=1}^{L}$ that chosen and tuned in a way to produces the closest output to $y \in \mathcal{Y}$ for input $x \in \mathcal{X}$ in dataset $\mathcal{D} = (x_{i}, y_{i})_{i=1}^{N}$. These weights and biases $w$ are fixed parameters optimized for best performance in a traditional neural network. However, the Bayesian approach states that we can place a distribution over the weights and biases instead of treating them as single variables. By placing a prior distribution $p(w)$ over the network's weights, we can introduce the likelihood function $p(y\mid x,w)$ that outputs the probability of $y$ given an input $x$. Such models are known as \textit{Bayesian neural networks} \cite{neal2012bayesian}.

Moreover, the idea of Bayesian neural networks is that by placing a distribution over weights instead of a single value, we will get a variety of outputs for input $x$ instead of the same output every time. Therefore, this will empower us to understand the model's behavior better. Since we are using the same input, if the variety of outputs is similar, we can conclude that the model has a high understanding and confidence for input $x$. On the contrary, if the model's output is different every time and the variety of outputs is not similar, we can conclude that the model has poor knowledge and confidence on input $x$. In the following sections, we will discuss how to use this understanding further to measure the model's uncertainty and utilize it to improve the model. 

For the problem of approximating complex probability distribution, a general approach is the variational inference, which states approximating the complicated posterior $p(w\vert\mathcal{D})$ with a simpler approximated distribution $q_{\theta}(w)$. Jensen’s inequality to the Kullback-Leibler divergence between the approximating distribution and the true posterior allows us to obtain the log-evidence lower bound $\mathcal{L}_{VI}$ \cite{smith2018understanding}.

\begin{equation}
		\mathcal{L}_{VI} = \int q_{\theta}(w)\log p(\mathcal{D}\vert w) \, \mathrm{d}w - D_{KL}(q_{\theta}(w)\vert\vert p(w))
\label{kb-div}
\end{equation}

\textcolor{black}{In Equation \ref{kb-div}, $p(w)$ is the prior distribution, $p(w\vert\mathcal{D})$ is the posterior distribution, and $q_{\theta}(w)$ is the approximated distribution.} Maximizing $\mathcal{L}_{VI}$ concerning $\theta$ will minimize the KL divergence between $q_{\theta}(w)$ and the model posterior, because the model evidence is a constant that does not depend on the parameters of $q_{\theta}$. The critical benefit is that the integration problem has been replaced with the optimization problem of maximizing parameterized function, which gradient descent methods can compute.

In the neural networks paradigm, a widely used tool for distribution approximation is dropout \cite{srivastava2014dropout}. The task of the dropout layer is to randomly set the output of neurons in its corresponding layer to zero. In other words, the dropout layer randomly disables part of the network to prevent overfitting and allow variations in overall output. \textcolor{black}{In this research, we use Monte Carlo Dropouts as an instance of Bayesian inference to create a variety in outputs in order to better understand the model's behavior}

The dropout layer probability, which is the probability that the output of a single neuron is set zero, is often assumed to be a constant and not a variable in this framework, but it is also possible to learn this parameter \cite{gal2017concrete}. Moreover, with the help of variational inference, the expectation over the posterior distribution can be estimated using the approximating distribution instead of the true posterior. The dropout distribution is still challenging to marginalize, but it is readily sampled from, so expectations can be approximated using the Monte Carlo estimator \cite{smith2018understanding}.

\begin{equation}
	\begin{aligned}
		E_{p(w\vert\mathcal{D})}[f^{w}(x)] &= \int p(w\vert\mathcal{D}) f^{w}(x) \, \mathrm{d}w \\
		&\simeq \int q_{\theta}(w) f^{w}(x) \, \mathrm{d}w \\
		&\simeq \frac{1}{T} \sum_{i=1}^{T}f^{w_{i}}(x),  w_{1..T} \sim q_{\theta}(w)
	\end{aligned}
	\label{mc-est}
\end{equation}

\subsection{Generative Active Learning}

The concept of generative active learning was first introduced in \cite{zhu2017generative} that utilized a Generative Adversarial Networks (GAN) to generate informative samples for the active learning cycle. Generating new samples needs groundwork like training GAN beforehand instead of just selecting samples from an unlabeled data pool. Therefore it leads to more complex and more extensive run time.

Generative base active learning framework continued in different directions. For instance, \cite{mayer2020adversarial} utilized GANs to generate informative samples but instead of using them directly in the query, found the closest sample in the unlabeled data pool to be labeled by an oracle and be added to the query. Moreover, \cite{tran2019bayesian} proposed that instead of generating a sample for labeling, the selected sample from the data pool can be labeled by an oracle, and then by using GAN, it can be transformed into a more informative sample, And both the base sample and the synthesized sample can be added to training dataset for retraining.

\subsection{Measures of Uncertainty}
\label{sec:2c}

We can break model uncertainty into two parts: aleatoric uncertainty (or statistical uncertainty) and epistemic uncertainty (or systematic uncertainty) \cite{der2009aleatory, hullermeier2021aleatoric, kendall2017uncertainties}.

1. \textit{Aleatoric}: This type of uncertainty represents inherent uncertainty in our data, which originates from intrinsic noise and variations in data points and data sources. Therefore, despite how much data we provide for the model, we would still have aleatoric uncertainty, so this type of uncertainty refers to as the irreducible part.

2. \textit{Epistemic}: this type of uncertainty represents the model's lack of knowledge and understanding. In simple terms, the model is not certain about an input because it has not seen and is not familiar with this variety of input. We can reduce this by providing more data for the model, so this type of uncertainty refers to as the reducible part.

Existing research on neural networks uncertainty introduced different measures of uncertainty. For example, \cite{li2017dropout} utilized expected entropy as a measure of uncertainty, and  \cite{rawat2017adversarial} used mutual information as the measure, whereas \cite{feinman2017detecting} estimated the variance of multiple draws from the predictive distribution. However, as argued in \cite{smith2018understanding}, entropy as a measure of uncertainty by itself does not distinguish between epistemic and aleatoric uncertainty. An alternative measure of uncertainty that can distinguish epistemic from aleatoric uncertainty is mutual information (MI) in the form \textcolor{black}{of Equation \eqref{mi_2}}. If we receive a new data point $x$ with a label $y$ for a model with weights $w$ that trained with dataset $\mathcal{D}$, the amount of information we would gain about the model parameters is given by:

\begin{equation}
	\begin{aligned}
		I(w;y\vert\mathcal{D},x) &= H[p(y\vert\mathcal{D},x)] - E_{p(w\vert\mathcal{D},x)}[H[p(y\vert w,\mathcal{D},x)]]\\
		& = H[p(y\vert\mathcal{D},x)] - E_{p(w\vert\mathcal{D})}[H[p(y\vert w,x)]]
	\end{aligned}
	\label{mi_2}
\end{equation}

Equation \eqref{mi_2} called Bayesian Active Learning by Disagreement (BALD) \cite{houlsby2011bayesian} and shows that if the model is already familiar with input $x$, it will gain little information by obtaining the label. In contrast, if the model is uncertain about input $x$ by knowing its label, it will gain information and become less uncertain. Therefore, the MI is a measure of uncertainty that captures \textit{epistemic} uncertainty. The separation between these types of uncertainty has been emphasised in the active learning paradigm \cite{nguyen2022measure}. Moreover, there are other applications that utilize the quantification of uncertainty. For example, \cite{lakshminarayanan2017simple} uses uncertainty estimations for out-of-distribution classification, and \cite{kendall2017uncertainties} investigates the effect of the increasing training set size on decreasing epistemic uncertainty. Also, \cite{wirges2019capturing} reviews how the uncertainty level depends on the object's distance regression task. \cite{sicking2022wasserstein}

\begin{figure}[t]
	\centering
	\includegraphics[width=0.48\textwidth]{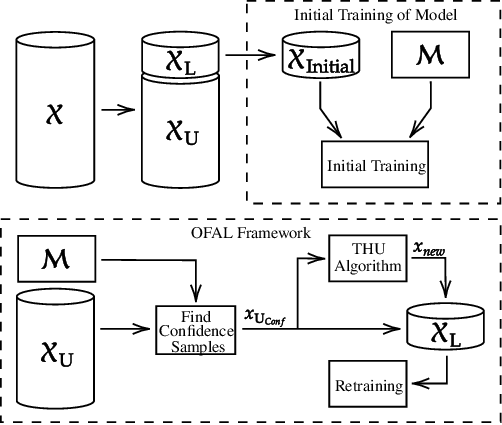}
	\caption{An overview of OFAL framework}
	\label{fig:framework}
\end{figure}

\section{Oracle-Free Active Learning Framework}

\label{sec:3}

We start by laying out the required structure and parameters. We have a large dataset $\mathcal{X}$ with labels $\mathcal{Y}$ formed from two parts: a smaller part $\mathcal{X}_L$ with its corresponding labels $\mathcal{Y}_L$ that has been labeled before, and a more prominent part $\mathcal{X}_U$ with its true labels $\mathcal{Y}_U$ that is unlabeled (note that $\mathcal{X}_L \cup \mathcal{X}_U = \mathcal{X}$ and $\mathcal{Y}_L \cup \mathcal{Y}_U = \mathcal{Y}$). The neural network model $\mathcal{M}$ has been trained initially with the $\mathcal{X}_L$ and has an accuracy of $\mathcal{A}$. Our task is to increase $\mathcal{A}$ by utilizing $\mathcal{X}_U$ without using an oracle to provide us the true labels.

To approach this task, we use a semi-active learning scheme. In traditional active learning, an oracle is responsible for finding the proper sample and labeling it. So after initial training of the model on the small labeled dataset $\mathcal{X}_{L}$ (also called $\mathcal{X}_{Initial}$), in every active learning query, the oracle looks through the unlabeled pool of data $\mathcal{X}_{U}$ and finds the sample that the model gains the most information by labeling it. Then oracle labels that sample and passes it to the model for learning.

By contrast, we introduce our semi-active learning scheme that does not need an oracle to label the new data. After the initial training on $\mathcal{X}_{L}$ dataset, we look through $\mathcal{X}_{U}$ to find samples with high prediction confidence from the model. These high confidence samples $\mathcal{X}_{U_{Conf}}$ are unlabeled, but the model predicted labels for them are probably correct. Nevertheless, since the model has high confidence in predicting these samples, we will not gain much information to improve the model solely by labeling them.

\begin{equation}
	\begin{aligned}
		\forall \; x \in \mathcal{X}_{U} \, :
		if \; Conf(\mathcal{M}(x_{U})) > T_{Conf} \qquad \qquad
		\\
		\qquad\longrightarrow \;
		y' \simeq y,
		\;  \;
		x \in \mathcal{X}_{U_{Conf}}
	\end{aligned}
	\label{conf}
\end{equation}

In Equation \eqref{conf}, $\mathcal{M}(\cdot)$ is the output of the model, $y$ is the true label for the sample $x$, $y'$ is the model's predicted label for the sample $x$, and $T_{Conf}$ is the threshold which if the sample confidence passes that will be counted as a confidence sample. As stated before, labeling these confidence samples would not help us improve the model, but the main idea of OFAL is that by transforming these samples to a more uncertain version of themselves, we would have a newly generated sample with a high uncertainty for the model. We can perform this transformation by increasing the uncertainty of the selected sample in respect to the model. Some examples of this transformation on selected samples are illustrated if Figure \ref{fig:trans}, and we will explain this task in detail in the following sections. Assuming that the model's predicted labels for confidence samples are correct, we add both the base confidence sample and the new uncertain versions to the initial training dataset and retrain the model \textcolor{black}{in the active learning framework. In other words, after the initial training of the model with the initial dataset, in each iteration, we add uncertain samples to our training set and train the model with the same weights as the previous iteration to improve the model's performance. We continue these iterations to obtain the required model performance.}

By adding these new samples to the training set and retraining the model for multiple queries, we expect that model to become more assertive in its uncertain areas and produce better and more robust results overall. The essential point is that we achieved this model improvement cost-free. Absent of an oracle means there is no cost for labeling data from an unlabeled pool, and we label these samples with the help of the model itself and then transform them into an instructive sample. Figure \ref{fig:framework} illustrates an overview of OFAL framework

\subsection{Uncertainty}
As asserted in the section \ref{sec:2c} and \cite{smith2018understanding}, we have a measure of uncertainty that captures epistemic uncertainty. While this quantity is not tractable analytically, it can be estimated by using the dropout approximation.

\bigskip
\begin{align}
	p(y\vert\mathcal{D},x) &\simeq
	\frac{1}{T}\displaystyle\sum_{i=1}^{T} p(y\vert w_i,x) 
	\label{mc}
	\\
	H[p(y\vert\mathcal{D},x)] &\simeq  H[\frac{1}{T}\displaystyle\sum_{i=1}^{T}p(y\vert w_i,x)]
	\label{pred_ent}
	\\
	E_{p(w\vert\mathcal{D})}[H[p(y\vert w,x)]] &\simeq \frac{1}{T}\displaystyle\sum_{i=1}^{T}H[p(y\vert w_i,x)]
	\label{expe_ent}
	\\\label{mi_3}
	I(w;y\vert\mathcal{D},x)&\simeq H[\frac{1}{T}\displaystyle\sum_{i=1}^{T}p(y\vert w_i,x)]
	\\
	&\qquad\quad-\frac{1}{T}\displaystyle\sum_{i=1}^{T}
	H[p(y\vert w_i,x)]
	\nonumber
\end{align}
\bigskip

\textcolor{black}{In the Equations \eqref{mc}, \eqref{pred_ent}, \eqref{expe_ent}, and \eqref{mi_3},} $w_i \sim q(w\vert\mathcal{D})$ are samples from dropout distribution, and the format demonstrated above can be approximate using the Bayesian interpretation of dropout, in fact Equation \eqref{mc} shows the Monte Carlo approximation of $y$ given an input $x$ ($p_{MC}(y\vert x)$). The first term of Equation \eqref{mi_3} \textcolor{black}{(Equation \eqref{pred_ent})} is the entropy of the predictive distribution, and the second term \textcolor{black}{(Equation \eqref{expe_ent})} is the mean of the entropy of the predictions \cite{smith2018understanding}. Using Monte Carlo Dropout, we can predict the model's output for a single point $x$ multiple times and get a variety of different outcomes. These outcomes could be closely similar if the model has a good understanding of $x$ or far different if the model is uncertain about $x$. Then we can calculate \eqref{mi_3} simply by calculating the entropy of mean of these multiple outputs for the first part of the equation and calculating the mean of entropy of them for the second part.

\begin{figure*}
	\centering
	\begin{subfigure}{0.45\textwidth}
		\includegraphics[width=\textwidth]{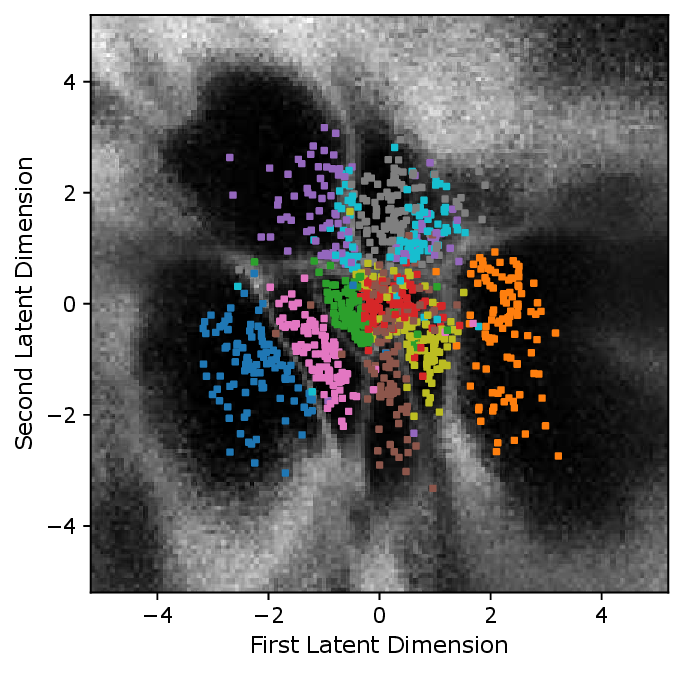}
		\caption{}
		\label{fig:latent_dim-a}
	\end{subfigure}
	\begin{subfigure}{0.45\textwidth}
		\includegraphics[width=\textwidth]{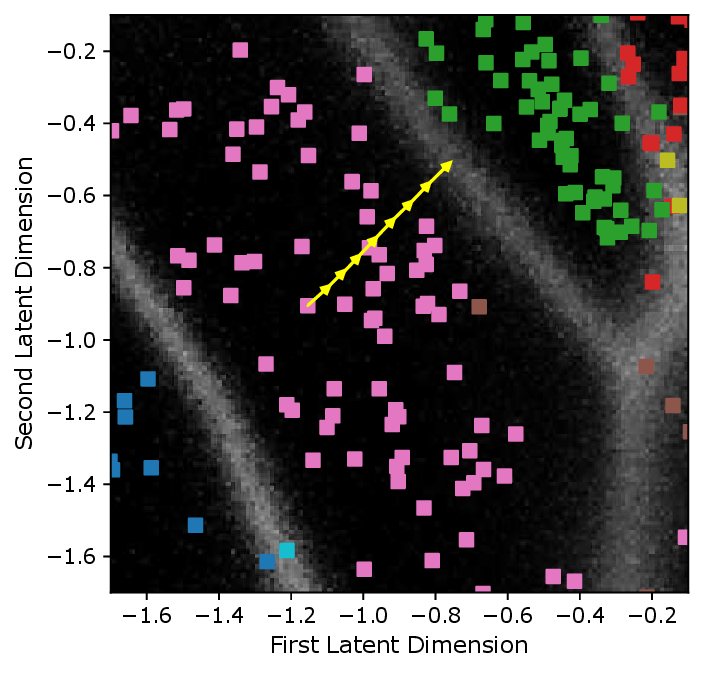}
		\caption{}
		\label{fig:latent_dim-b}
	\end{subfigure}
	\\
	\begin{subfigure}{0.45\textwidth}
		\includegraphics[width=\textwidth]{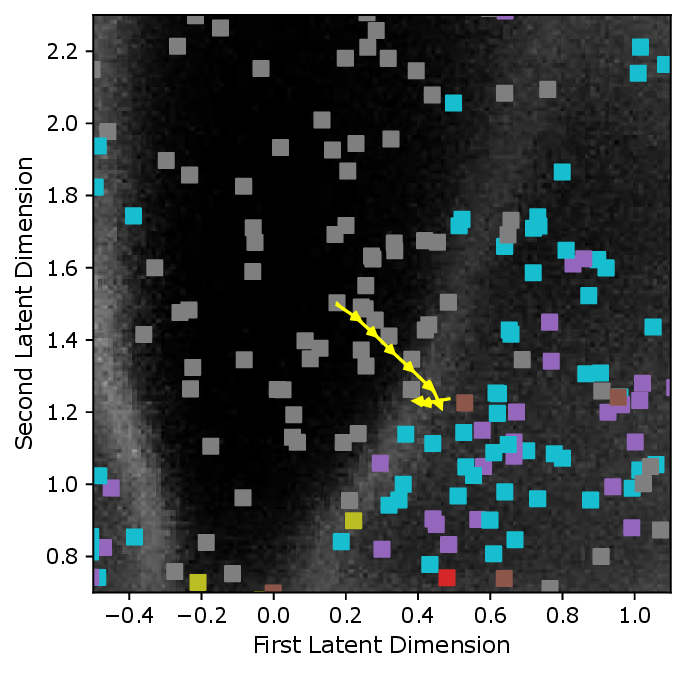}
		\caption{}
		\label{fig:latent_dim-c}
	\end{subfigure}
	\begin{subfigure}{0.45\textwidth}
		\includegraphics[width=\textwidth]{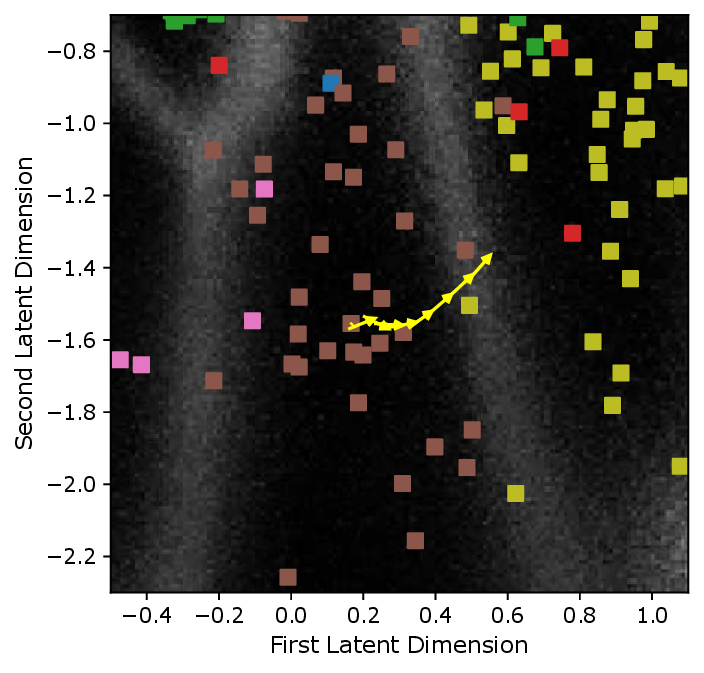}
		\caption{}
		\label{fig:latent_dim-d}
	\end{subfigure}
	\caption{
		(a) Uncertainty of the model illustrated in latent space of 2-dimensional VAE, different colors represent classes in initial dataset, and the lighter background represents higher uncertainty. (b), (c), and (d) Illustrates THU algorithm for samples. THU starts with a confidence sample (shown with darker background) and steps toward a more uncertain part of the latent dimension (shown with lighter background) to generate a new uncertain sample.
	}
	\label{fig:latent_dim}
\end{figure*}

\subsection{Toward Higher Uncertainty}

Now that we have a measure of uncertainty and a method to quantify it, we can proceed to the next step. As we argued, if we use model uncertainty to find more uncertain samples for the model, these newly founded samples are highly informative. First, we introduce loss function $\mathcal{L}$ that formed from two parts. The former part of $\mathcal{L}$ is the uncertainty that we want to maximize, and the latter part is a restriction that helps to prevent the newly generated sample $x_{new}$ from differing too much from its original form $x_{old}$.

\begin{equation}
	\mathcal{L} = - Uncertainty + \alpha \times Restriction
	\label{loss_1}
\end{equation}

In Equation \eqref{loss_1}, $\alpha$ is a constant that acts as a weight for the restriction part of $\mathcal{L}$ to bring balance between former and latter part. Note that the restriction can be any form of distance base criteria that measures the difference between $x_{old}$ and $x_{new}$ (such as Euclidean distance, Mean square error).

In traditional neural network training, its weights are variables trying to find the best output for a set of fixed inputs. In this paper, we reverse this standard by assuming the model's weights as a set of fixed constants and approach inputs as variables that can be tuned to minimize the loss $\mathcal{L}$. In other words, we input $x$ to a pre-trained model with weights $w$ to find the model's output, and with the help of Dropout approximation and Equation \eqref{mi_3}, we can calculate $\mathcal{L}$. Then by using Gradient descent, we try to modify $x$ in a way that minimizes $\mathcal{L}$. Note that the model's weights $w$ do not change throughout this method and are the same $w$ as the start, and the product of this method is a new form of $x$ called $x_{new}$ that model is more uncertain regarding it and even may cause erroneous behavior from the model.

\begin{figure*}
	\centering
	\begin{subfigure}{\textwidth}
		\includegraphics[width=\textwidth]{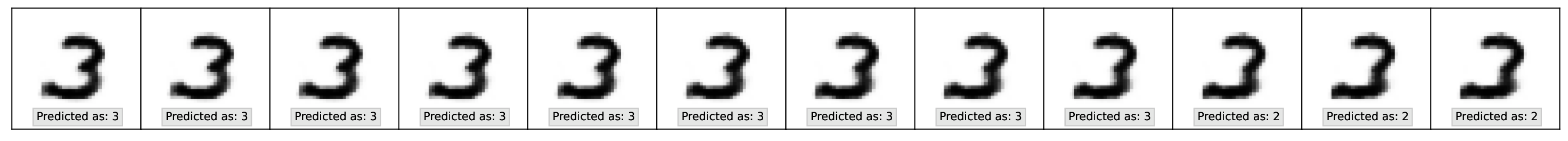}
		\caption{}
		\label{fig:numbers-a1}
	\end{subfigure}
	\\
	\begin{subfigure}{\textwidth}
		\includegraphics[width=\textwidth]{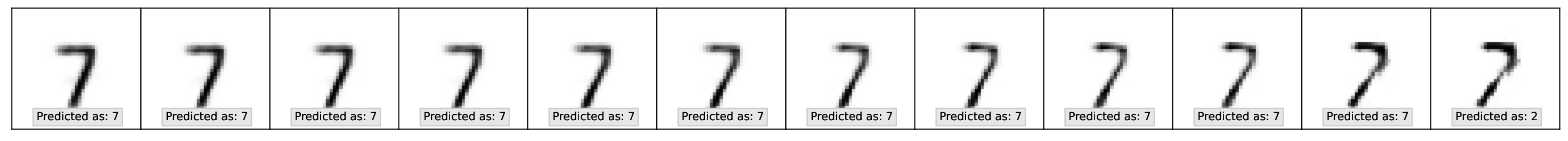}
		\caption{}
		\label{fig:numbers-a2}
	\end{subfigure}
	\\
	\begin{subfigure}{\textwidth}
		\includegraphics[width=\textwidth]{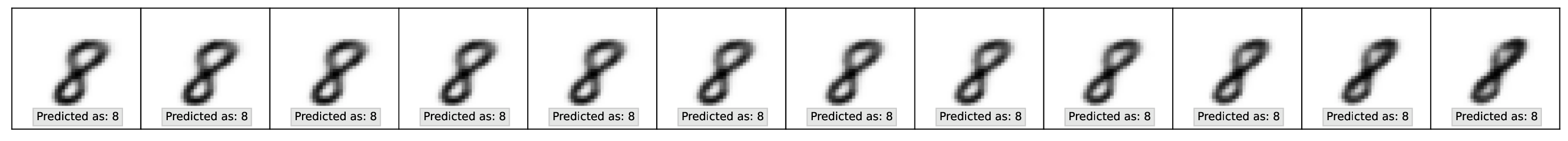}
		\caption{}
		\label{fig:numbers-a4}
	\end{subfigure}
	\\
	\begin{subfigure}{\textwidth}
		\includegraphics[width=\textwidth]{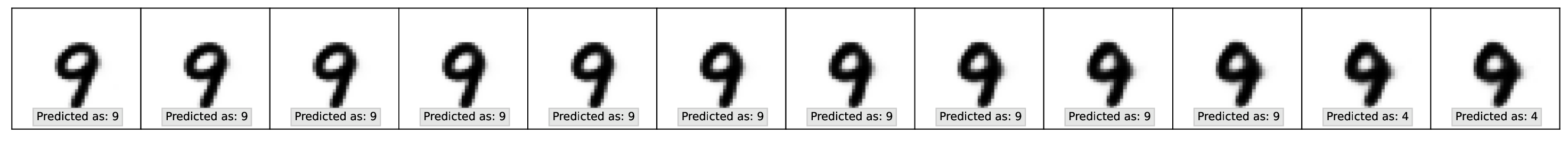}
		\caption{}
		\label{fig:numbers-b}
	\end{subfigure}
	\\
	\begin{subfigure}{\textwidth}
		\includegraphics[width=\textwidth]{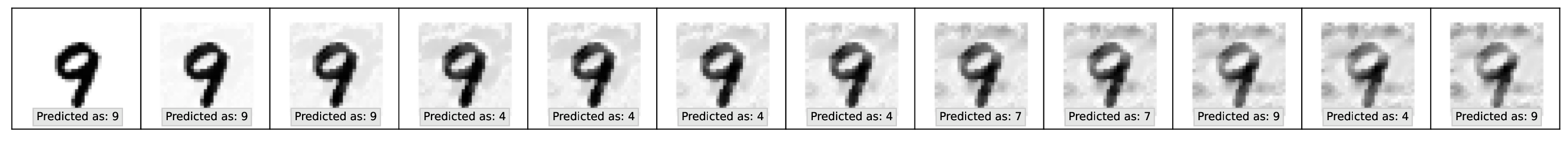}
		\caption{}
		\label{fig:numbers-c}
	\end{subfigure}
	
	\caption{
		In all sections, we started with a random sample from the model's training dataset (far-left sample) and moved toward samples with higher uncertainty (far-right sample). (a) trough (d) are utilizing a VAE, and (e) has the same starting sample as (d), but it is not using a VAE.
	}
	\label{fig:numbers}
\end{figure*}

\subsection{Variational Autoencoder}

So far, we have described the measure of uncertainty and steps to find more uncertain samples, but this newly generated more uncertain sample $x_{new}$ may not be meaningful by itself. Neural network inputs usually have too many degrees of freedom, and in the process of transforming this input to find a more uncertain one, we may generate samples that are not relevant concerning our datasets and classes or outputs. Therefore we need an approach to decrease this freedom to step toward meaningful and more uncertain samples.

In order to address this concern, we utilize a variational autoencoder (VAE) \cite{kingma2013auto}. Encoding all of the data creates a space where every point is meaningful in relation to our data, so we chose the variable being affected by the gradient descent to be a point in VAE's latent space. Thus, after encoding an input $x$ to its corresponding point $z$ in latent space, it can be decoded again to $x$ format to be inputted into the model, and we can change $z$ so that its decoded format be a more uncertain sample for the model. In other words, we start from an input $x$, encode it to find $z$, and walk in the latent space created by VAE to eventually find a new point $z_{new}$ that can be decoded to $x_{new}$ which is our new uncertain sample.

\textcolor{black}{To better understand the transformation of samples and the generation of new and more uncertain samples, we try to follow their pass into VAE's latent space. For this, we use a 2-dimensional VAE (since we cannot plot a higher-dimensional VAE's latent space). This is illustrated in Figure \ref{fig:latent_dim}. Figure \ref{fig:latent_dim-a} shows described VAE's latent space, in which different colors represent classes in the dataset, and the lighter background represents higher uncertainty. In Figures \ref{fig:latent_dim-b}, \ref{fig:latent_dim-c}, and \ref{fig:latent_dim-d}, we can see that a confidence sample has been selected as the seed sample for the algorithm, hence, it has darker background corresponding to the areas with lower uncertainty. As shown by the yellow arrows, we step toward more uncertain areas of latent space, which have lighter backgrounds corresponding to their higher uncertainty. These uncertain areas are usually located on the edges of the class distributions or where the borders of two or more classes encounter each other.}

\begin{figure*}
	\centering
	\includegraphics[width=0.75\textwidth]{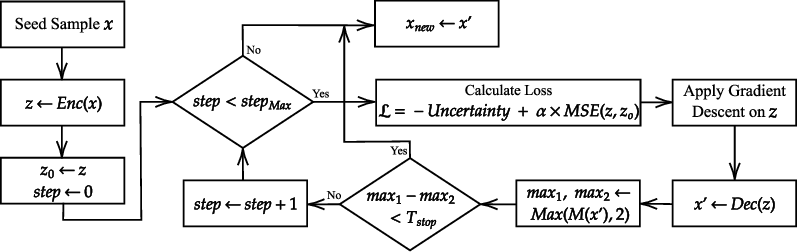}
	\caption{\textcolor{black}{THU algorithm's block diagram}}
	\label{fig:thu}
\end{figure*}

The presence and absence of VAE are illustrated in Figure \ref{fig:numbers}. In the absence of VAE, by applying gradient descent in Figure \ref{fig:numbers-c} instead of transforming the sample to a meaningful sample, it changes parts of the sample that are not important. Even though it generated an uncertain sample for the model, this new sample can not be a part of dataset $\mathcal{X}$. However, in \ref{fig:numbers-b} that started with the same sample, in the presence of VAE, it generated a new and more uncertain sample that causes erroneous behavior from the model. In a way, by doing this, we are stepping into areas of latent space that have higher uncertainty. These areas are usually at the border where different classes meet in latent space, and since we are measuring epistemic uncertainty instead of aleatoric uncertainty, these uncertain areas are not meaningless noise but meaningful samples that the model has not seen at all or has not seen enough before. Algorithm \ref{alg:thu}: Toward Higher Uncertainty (THU) illustrates the structure of finding these new samples. \textcolor{black}{A more detailed illustration of Algorithm \ref{alg:thu} could be found in Figure \ref{fig:thu}, in which we can follow the path of a sample $x$, which transforms into its coded version $z$ and then it goes through the THU algorithm to generate a new and more uncertain sample $x_{new}$.}

\begin{algorithm}
	\caption{Toward Higher Uncertainty (THU)}\label{alg:thu}
	\begin{algorithmic}[1]
		
		\algrenewcommand\algorithmicrequire{\textbf{Input:}}
		\Require
		$x\in\mathcal{X}$: Starting seed sample,
		$\mathcal{M}$: Model,
		$Enc$: VAE's Encoder, 
		$Dec$: VAE's Decoder
		\algrenewcommand\algorithmicrequire{\textbf{Output:}}
		\Require
		$x_{new}$: Newly generated uncertain sample
		\algrenewcommand\algorithmicrequire{\textbf{Constant:}}
		\Require
		$step_{Max}$: Maximum steps for finding uncertain sample,
		$T_{stop}$: Threshold to stop THU algorithm from going to far
		\State $z \gets Enc(x)$
		\State $z_{0} \gets z$
		\State $step \gets 0$
		
		\While{$step<step_{Max}$}
		\State $loss \gets - Uncertainty(Dec(z)) + MSE(z,z_{0})$\
		\State $grad \gets apply\_ gradient \_ descent(loss)$
		\State $z \gets z + grad$
		\State $x' \gets Dec(z)$
		\State $max_1, max_2\gets Max(M(x'),2)$
		\If{$max_1 - max_2 < T_{stop}$}
		\State break
		\EndIf
		\State $step \gets step + 1$
		\EndWhile
		\State $x_{new}\gets x'$
		\State \Return{$x_{new}$}
	\end{algorithmic}
\end{algorithm}
\bigskip

\begin{figure*}
	\centering
	\begin{subfigure}{0.60\textwidth}
		\includegraphics[width=\textwidth]{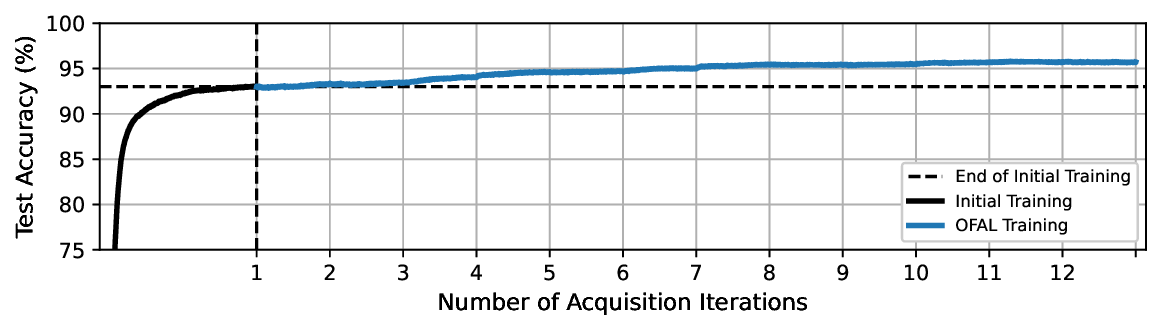}
		\caption{Model accuracy on test set}
		\label{fig:bald-a}
	\end{subfigure}
	\\
	\begin{subfigure}{0.60\textwidth}
		\includegraphics[width=\textwidth]{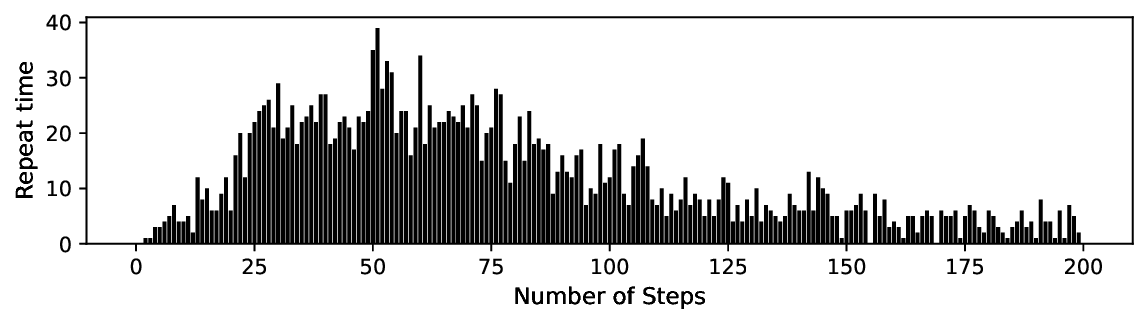}
		\caption{Number of steps taken to generate \textcolor{black}{uncertain} samples}
		\label{fig:bald-b}
	\end{subfigure}
	\caption{
		The model reached 93.00\% accuracy on the test dataset following the initial training. After 12 iterations (in each iteration, 400 samples were added each time to the training set, containing 200 samples from the unlabeled data pool), the model reached 95.70\% accuracy on the test dataset.
	}
	\label{fig:bald}
\end{figure*}

\section{Empirical Evaluation}
\label{sec:4}

In this section, we demonstrate the effectiveness of our proposed method. For this matter, we use MNIST \cite{lecun1998gradient} 28$\times$28 handwritten digits dataset (60000 training samples, 10000 testing samples, in 10 classes). We randomly select 1000 samples (equal in each class) from MNIST's training set as labeled data for initial training and consider the rest as unlabeled data pool. Then we overtrain the model on the initial training set to learn every bit of information that it can from the initial set; therefore, any improvement afterward would be because of our proposed scheme. Note that the model cannot overfit the initial training set because of the dropout layers, so overtraining the model would not cause issues. Also, the VAE will be trained on all $\mathcal{X}$ ($\mathcal{X}=\mathcal{X}_{L}\cup\mathcal{X}_{U}$) dataset since it does not need any labels and solely transfer all the data set $\mathcal{X}$ to a latent dimension.

\subsection{Improving Model with OFAL}
\label{sec:4-1}

After the initial training of the model is done, OFAL begins its task to improve the model. As presented before, we look toward the unlabeled data pool and find samples with high confidence. In each iteration, we assume model output for any samples with at least 99\% prediction confidence from the model ($T_{conf}$ = 0.99) is correct, and from these confidence samples, we randomly select 200 samples (20 for each class of MNIST 10 classes) called $\mathcal{X}_{conf}$ with labels $\mathcal{Y}_{conf}$. Next, with the help of the Algorithm \ref{alg:thu}, we increase uncertainty for all 200 chosen samples and call them $\mathcal{X}_{uncr}$ and use their baseline labels for them. For generating new samples, we chose stopping criteria of $T_{stop}$=0.4, which means if the difference between the probability of predicted class (that previously was 99\% or above and will be decreasing with each step) and probability of second highest predicted class (which will be increasing after each step) \textcolor{black}{become lower than} $T_{stop}$, we reached our uncertain sample.

\begin{table}
	
	\caption{Accuracy of the neural network model at different stages}
	\centering
	\setlength{\tabcolsep}{3pt}
	\begin{tabular}{p{100pt} p{100pt} p{0pt}}
		\toprule
		\centering Neural Network Model & \centering Accuracy on Test Set &\\
		\midrule
		\centering Model after Initial Training &  \centering 93.00\% &\\
		
		\centering Model after OFAL Method &  \centering 95.70\% &\\
		\bottomrule
	\end{tabular}
	\label{tab:1}
\end{table}

\begin{figure*}
	\centering
	\begin{subfigure}{0.40\textwidth}
		\includegraphics[width=\textwidth]{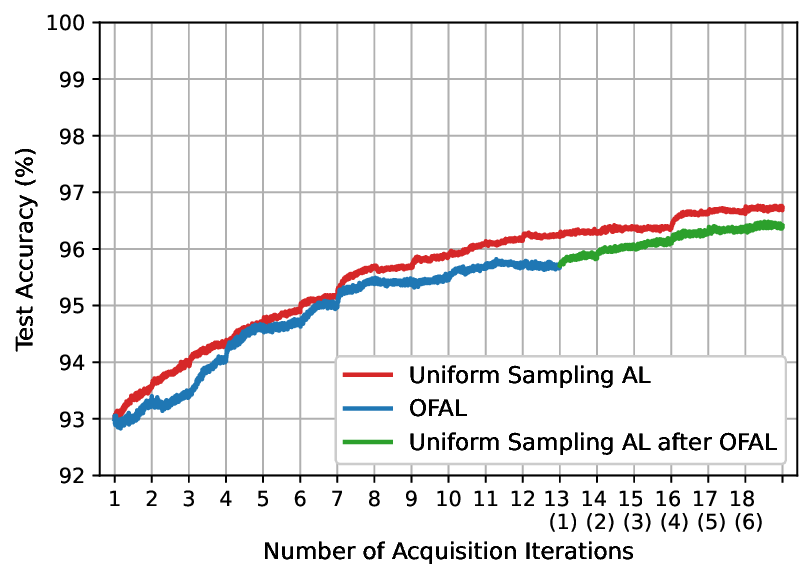}
		\caption{Uniform Sampling}
		\label{fig:sm_after_bald-a}
	\end{subfigure}
	\begin{subfigure}{0.40\textwidth}
		\includegraphics[width=\textwidth]{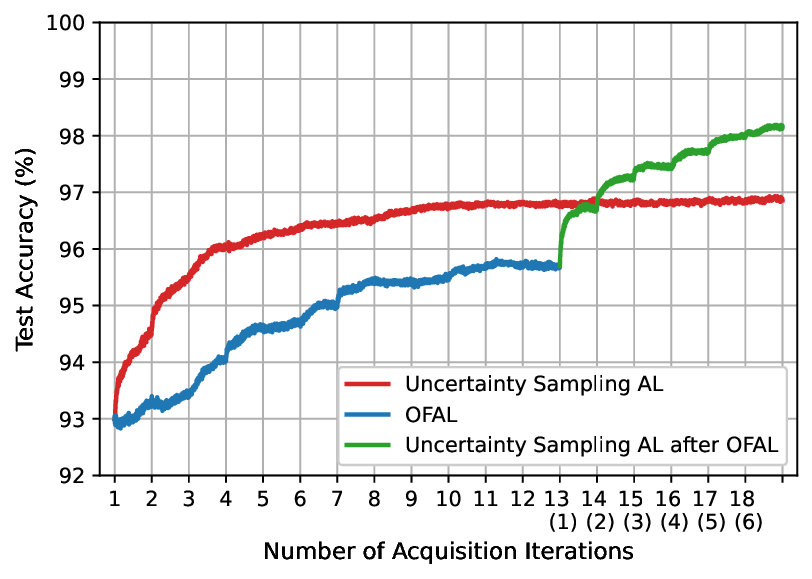}
		\caption{Uncertainty Sampling}
		\label{fig:sm_after_bald-b}
	\end{subfigure}
	\caption{
		(a) OFAL performs similar to uniform sampling and reaches the same results without labeling new data. (b) Utilizing  uncertainty sampling after OFAL reaches the same accuracy as solo uncertainty sampling with one-tenth of labeled samples.
	}
	\label{fig:sm_after_bald}
\end{figure*}

After generating new samples, we add both $(\mathcal{X}_{cont},\mathcal{Y}_{conf})$ and $(\mathcal{X}_{uncr},\mathcal{Y}_{conf})$ to model's training set to form a new query. So in every acquisition iteration, 400 samples will be added to the training set, 200 samples from selecting and labeling samples from the unlabeled data pool, and 200 samples from newly generated uncertain samples. We acquire samples and retrain the model for multiple iterations; results are illustrated in Figure \ref{fig:bald}. \textcolor{black}{In Figure 6a, it is shown that the model, after the initial training, reached 93.00\% accuracy on the test dataset (demonstrated by the black line). At this point, the model is overtrained so that it can achieve the highest performance that the model can reach on the initial dataset. The dashed lines show the end of the initial training after 10000 epochs and the accuracy achieved by the model after the initial training. Moreover, using the OFAL framework shows that the model's accuracy on the test dataset is increasing (demonstrated by the blue line). After 12 iterations, the model's accuracy is enhanced to 95.70\% on the test dataset. This improvement was caused by utilizing active learning without an expensive oracle to label the selected data from the unlabeled data pool.} Table \ref{tab:1} shows the end result.

Note that after applying the THU algorithm for a sample, if the steps that have been taken to reach the stopping criteria matches the maximum steps allowed ($step_{max}$), we assume that the route to find the required uncertain version of this sample has stopped by a local extremum. Since this local extremum can be located close to the seed sample and the seed sample is a confidence sample, this newly generated sample may not contain much information to improve the model. So instead of waisting this prospect, we set aside this sample and randomly chose another confidence sample from the confidence samples with the same class label. Likewise, if a sample reaches its uncertain version with zero steps taken, we will set aside that sample and randomly select another one (This can acquire if a lower $T_{conf}$ is chosen).

\textcolor{black}{There are two hyper-parameters that the process of choosing their value needs further explanation: Confidence threshold $T_{conf}$ and stopping threshold $T_{stop}$. For confidence threshold $T_{conf}$, two criteria need consideration. First, the $T_{conf}$ should be as high as possible to act as a strong filter, so that models predicted labels for all selected samples would be correct. Secondly, the $T_{conf}$ should be low enough so that a fair amount of samples from all classes go through the system. In our experience in the implementation, our first choice was $T_{conf}=99.99\%$. Although the predicted label for all samples selected with $T_{conf}=99.99\%$ was correct, the quantity of these samples was not enough to select the needed amount of samples in every class. After experiments with several values for this parameter, we chose $T_{conf}=99\%$. With $T_{conf}=99\%$, we have enough samples to proceed with the implementation, and the predicted labels for chosen samples have almost 100\% accuracy in any iteration.}

\textcolor{black}{Next is stopping threshold $T_{stop}$. For this, we first experimented with generating a new sample, as uncertain as possible. This means choosing $T_{stop}=0$. To better understand this, we can utilize the latent space of VAE illustrated in Figure \ref{fig:latent_dim}. Choosing $T_{stop}=0$ means getting significantly close to the border of the class and where two different classes meet each other. These ultimately result in the most uncertain samples, but these samples are not necessarily considered good samples for learning. By getting too close to the border, we generate samples with features of both classes simultaneously. These samples can be classified as one class by a person and classified as another class by another person. Hence, these can not be reliable samples for the learning process of the model. Therefore, our goal would be to get close to the borders, where most uncertain areas of latent space are located, without generating samples that cannot be undoubtedly classified into one of the classes. By experimenting with different values for $T_{stop}$ and observing the generated samples and their usefulness in improving the model's performance, we concluded that $T_{stop}=0.4$ to be most suitable value to get the finest outcome.}

\begin{figure*}
	\centering
	\begin{subfigure}{0.40\textwidth}
		\includegraphics[width=\textwidth]{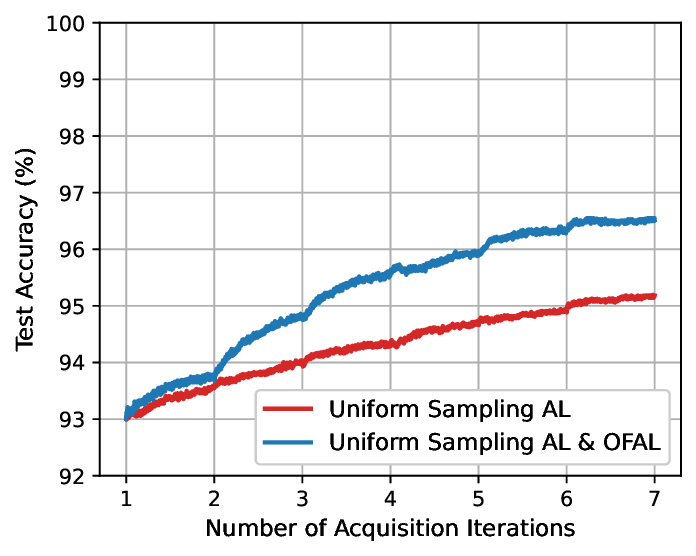}
		\caption{Uniform Sampling}
		\label{fig:sm_with_bald-a}
	\end{subfigure}
	\begin{subfigure}{0.40\textwidth}
		\includegraphics[width=\textwidth]{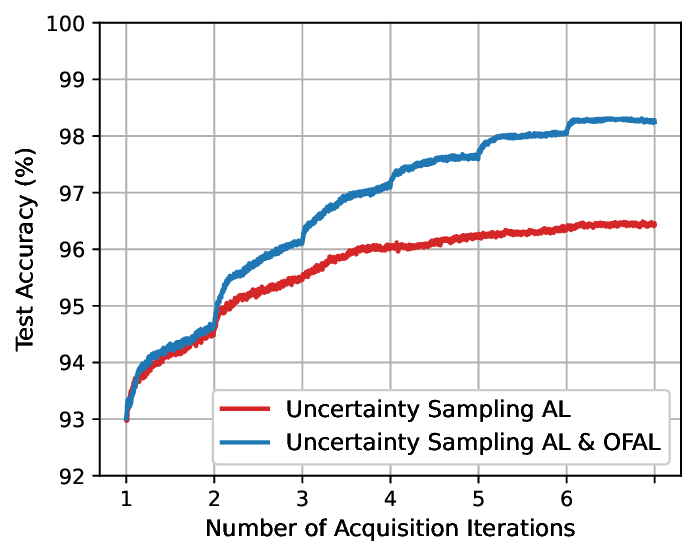}
		\caption{Uncertainty Sampling}
		\label{fig:sm_with_bald-b}
	\end{subfigure}
	\\
	\begin{subfigure}{0.40\textwidth}
		\includegraphics[width=\textwidth]{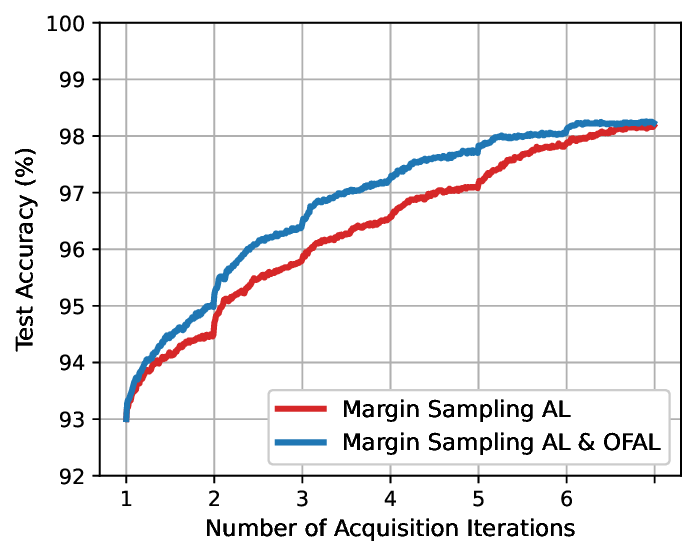}
		\caption{Margin Sampling}
		\label{fig:sm_with_bald-c}
	\end{subfigure}
	\begin{subfigure}{0.40\textwidth}
		\includegraphics[width=\textwidth]{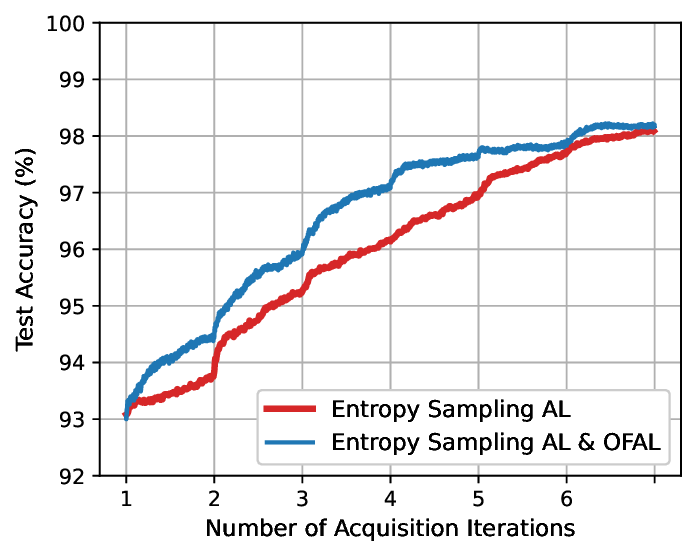}
		\caption{Entropy Sampling}
		\label{fig:sm_with_bald-d}
	\end{subfigure}
	\caption{
		By integrating OFAL with different sampling methods, we can reach better results with fewer labeled data. 
	}
	\label{fig:sm_with_bald}
\end{figure*}

\subsection{Comparison with Other Sampling Methods}

To witness the performance of OFAL, we implement two widely used sampling methods: uniform sampling (random sampling) and uncertainty sampling (unrelated to our method, selects samples with the least prediction probability). Since all of these sampling methods need an oracle to label their chosen unlabeled samples, the comparison is not entirely fair, even though our method exhibited excellent results spite not needing an oracle. Hence, after the initial training of the model, we use each sampling method for multiple iterations to retrain the model. In each acquisition iteration, the sampling method selects 200 samples from the unlabeled data pool and transfers them to the oracle for labeling, and newly labeled samples form a query to be added to the training set for the model's retraining. However, for our proposed method,  after initial training and the improvement of baseline model with uncertain samples (which we discussed in Section \ref{sec:4-1} and Figure \ref{fig:bald}), we use the same sampling method in the same manner on top of our improved model.

Figure \ref{fig:sm_after_bald} illustrated the result. We can observe what sampling methods achieved with a substantial amount of acquisition iterations, using OFAL before sampling methods accomplished the same result with significantly less labeled data. For example, in Figure \ref{fig:sm_after_bald-b} we can notice the solo uncertainty sampling after ten acquisition iterations, and the cost of labeling 2000 unlabeled samples from the data pool reached the same result that using uncertainty sampling after OFAL with one acquisition iterations and cost of labeling 200 unlabeled samples. Additionally, the overall improvement of the model by using our method before uncertainty sampling is remarkably higher than using solo uncertainty sampling, which reached its upper limits after ten iterations.

\subsection{Integration with Other Sampling Methods}

Another approach to utilizing our method is integrating it with other sampling methods. Because of its cost-free nature by not needing an oracle, it can be used simultaneously as other sampling methods, meaning when other sampling methods do their sample selection and labeling, we generate and label our samples simultaneously and add them to the same query for the model's retaining. We use two established sampling methods for evaluation: Magin sampling and Entropy sampling. After the model's initial training in each iteration, the sampling method selects 200 samples from the unlabeled data pool and passes them to the oracle for labeling. Then we use our method to select 200 confidence samples data pool with their predicted labels and pass them to generate uncertain samples. Finally, after each acquisition iteration, we form a 600 samples query containing 200 samples from the sampling method (which their labels provided by the oracle), 200 confidence samples (which we assumed model's predicted label for them is correct since they have high prediction confidence), and 200 generated uncertain samples (which we use the same label as their base model).

Results are presented in Figure \ref{fig:sm_with_bald}. We see that by utilizing OFAL, we can enhance every other sampling method that needs an oracle. Therefore, OFAL enables us to reach greater results with fewer labeled samples or even more significant overall results.

\section{Conclusion} 
\textcolor{black}{This paper introduces OFAL, an oracle-free active learning scheme that can improve neural networks' performance without an expensive oracle to label new data from an unlabeled data pool. We started by discussing the idea of Bayesian neural networks and the variety in models output that can help us better understand the model's performance. We implement this by using Monte Carlo dropouts. Then we quantify epistemic uncertainty by utilizing the Monti Carlo estimation and the variety of outputs created with the dropout layer's existence. Then, we generated new and more uncertain samples concerning the model and utilized a VAE to keep these newly generated samples meaningful. We illustrated that we walk through the VAE's latent space toward more uncertain areas and decode them to find more uncertain samples. We discussed that since these samples are more uncertain with respect to our model, learning them would be highly beneficial to improve the model's performance. Finally, we form an active learning query to retain the model. We active $2.7\%$ improvement on the accuracy of a CNN trained on a small part of the MNIST dataset. Additionally, we compared and integrated OFAL with other widely used oracle sampling methods, and we were able to reach better accuracy with less labeled samples or reach better outcomes overall. Since the OFAL framework dose does not need additional labeling costs, it can work simultaneously with any active learning method and generate informative samples to improve the model's performance.}

\section*{Appendix}

\section*{\textcolor{black}{I. Neural Network Architecture \& Hyper Parameters}}\label{secA1}

\subsection{\textcolor{black}{Neural Network Architecture}}

This section describes the neural networks used in the development of this paper. There are two neural networks: First is the convolutional neural network (CNN) \cite{o2015introduction} which is the model we are trying to improve, and second is the variational autoencoder which is used to create the latent space.

\begin{figure}[H]
	\centering
	\setlength{\belowcaptionskip}{-8pt}
	\includegraphics[width=0.48\textwidth]{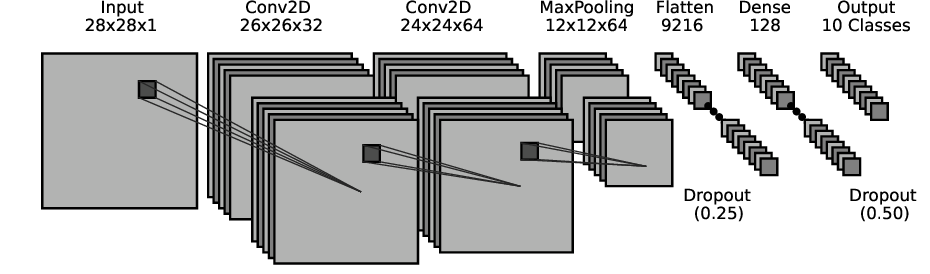}
	\caption{The architecture of the convolutional neural network model, which has been used as the model we are trying to improve.}
	\label{fig:cnn}
\end{figure}

For the model, we used a simple CNN for image classification. As it is illustrated in Figure \ref{fig:cnn}, the neural network is built with two 2D convolutional layers (with a kernel size of 3), one 2D max pooling layer (with a pool size of 2), flatten layer, one fully connected dense layer, and a 10-class final layer. ReLU activation is used for this network, and the last layer has a Softmax activation. There are 2 Dropout layers have been used to prevent the model from overfitting.

\begin{figure}[H]
	\centering
	\setlength{\belowcaptionskip}{-15pt}
	\includegraphics[width=0.48\textwidth]{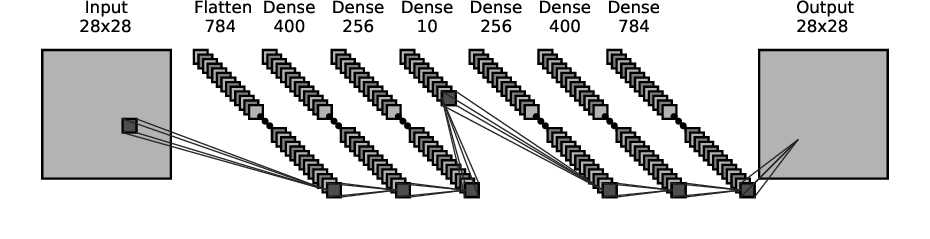}
	\caption{The architecture of the variational autoencoder, which has been used to create the latent space}
	\label{fig:vae}
\end{figure}

Next is the variational autoencoder. As shown in Figure \ref{fig:vae}, several fully connected dense layers have been used. For the encoder, flatten layer and two dense layers have been used before the bottleneck. In the decoder part, there are 3 dense layers before the reconstruction of the input image. For the latent space, we experimented with different numbers of dimensions ranging from 2 to 30. For the MNIST dataset, 10 latent space dimensions are both accurate and sufficient. For the illustration of Figure \ref{fig:latent_dim}, 2-dimensional latent space was used.

\bibliographystyle{abbrv}
\bibliography{template.bib}

\end{document}